\documentclass[lettersize,journal]{IEEEtran}

\usepackage{amssymb,amsmath}
\usepackage{cite}
\usepackage{graphicx}
\usepackage{subfig}

\usepackage{psfrag}
\usepackage{url}
\usepackage[latin1]{inputenc}
\usepackage[absolute,overlay]{textpos}
\usepackage{gensymb}
\usepackage{cases}
\usepackage[font=footnotesize]{caption}
\usepackage{float}
\usepackage[linesnumbered,ruled,lined]{algorithm2e}
\usepackage{pgf}

\usepackage[nocomma]{optidef}

\usepackage{amsthm}

\usepackage{booktabs}
\usepackage{color,soul}

\usepackage{textcomp}
\usepackage{bbm}
\usepackage{xcolor}
\def\BibTeX{{\rm B\kern-.05em{\sc i\kern-.025em b}\kern-.08em
    T\kern-.1667em\lower.7ex\hbox{E}\kern-.125emX}}

\usepackage{tabularx}
\usepackage{makecell}
\usepackage{multirow}

\begin{document}

\title{Resilient UAV Trajectory Planning via Few-Shot Meta-Offline Reinforcement Learning}
\author{
	\IEEEauthorblockN{Eslam Eldeeb and Hirley Alves
}
	
    \thanks{Eslam Eldeeb and Hirley Alves are with the Centre for Wireless Communications (CWC), University of Oulu, Finland. (e-mail: eslam.eldeeb@oulu.fi; hirley.alves@oulu.fi).
    }
    
    \thanks{This work was supported by 6G Flagship (Grant Number 369116) funded by the Research Council of Finland.}
}
\maketitle

\begin{abstract}

Reinforcement learning (RL) has been a promising essence in future 5G-beyond and 6G systems. Its main advantage lies in its robust model-free decision-making in complex and large-dimension wireless environments. However, most existing RL frameworks rely on online interaction with the environment, which might not be feasible due to safety and cost concerns. Another problem with online RL is the lack of scalability of the designed algorithm with dynamic or new environments. This work proposes a novel, resilient, few-shot meta-offline RL algorithm combining offline RL using conservative Q-learning (CQL) and meta-learning using model-agnostic meta-learning (MAML). The proposed algorithm can train RL models using static offline datasets without any online interaction with the environments. In addition, with the aid of MAML, the proposed model can be scaled up to new unseen environments. We showcase the proposed algorithm for optimizing an unmanned aerial vehicle (UAV) 's trajectory and scheduling policy to minimize the age-of-information (AoI) and transmission power of limited-power devices. Numerical results show that the proposed few-shot meta-offline RL algorithm converges faster than baseline schemes, such as deep Q-networks and CQL. In addition, it is the only algorithm that can achieve optimal joint AoI and transmission power using an offline dataset with few shots of data points and is resilient to network failures due to unprecedented environmental changes.

\end{abstract}

\section{Introduction}\label{sec:introduction}

\subsection{Context and Motivation}
Recent progress towards intelligent wireless networks embraces efficient and fast decision-making algorithms. Machine learning / artificial intelligence (ML/AI) has gained further interest in 5G-beyond and 6G systems due to their powerful decision-making algorithms that can adapt to large and complex wireless networks~\cite{10198239,wang2020artificial}. Reinforcement learning (RL) is one family of ML/AI that is known as the algorithm of decision-making. In RL, an agent observes the environment, makes decisions, and receives an award that evaluates how good the decision is in the current environment observation. A policy in RL describes what decisions to select at each observation. RL aims to find the optimum policy that maximizes the received awards~\cite{sutton2018reinforcement}.

To this end, RL has shown great promise in a wide range of applications, such as radio resource management (RRM)~\cite{6542770}, network slicing~\cite{8540003}, unmanned aerial vehicle (UAV)~\cite{eldeeb2022multi} networks, connected and autonomous vehicle (CAV) networks~\cite{9762548}. RL's power relies on RL algorithms' ability to handle model-free systems, where it is tough to formulate an efficient closed-form model to the system~\cite{8714026}. This applies to 5G-beyond and 6G systems, which are often very complex, have large dimensions, and are full of uncertainties. These characteristics of the wireless systems make RL algorithms fit most of the problems the wireless systems face~\cite{chen2021deep}. In addition, the breakthrough in deep RL enables solving extraordinarily complex and large systems by combining deep neural networks (DNNs) with traditional RL algorithms~\cite{DQNs}.

One example where RL and deep RL show superior benefits is autonomous UAVs' trajectory optimization. UAVs provide remarkable flexibility for gathering data from remote sensors and enhance communication by flying closer to the sensors, thereby increasing the likelihood of line-of-sight (LoS) communication~\cite{10587008}. Moreover, remote sensors, such as those used in smart agriculture networks, often have limited power supplies and are difficult to access for battery replacement, particularly during adverse weather conditions~\cite{9316211, RAJ2024100300, 10423798}. UAVs play a crucial role in conserving sensor power by reducing the distance between them. Scalable RL algorithms can optimize UAV trajectory and scheduling policies, even in dynamic and rapidly changing network environments~\cite{9086620,9882524}.

Despite its high applicability to the wireless environment, RL and deep RL still face significant difficulties in real-world wireless systems~\cite{levine2020offline}. First, almost all RL and deep RL algorithms designed for wireless communication applications are online. Online RL counts on continuous interaction with the environment to update the learned policies until converging to the optimum policy. However, online interactions might not be feasible in real-world scenarios. For instance, online interactions might be unsafe in some applications, such as UAV and CAV networks, where bad decisions can lead to hazardous consequences. In addition, online interactions might be costly and time-consuming in some applications, such as RRM and network slicing, where the algorithm spends large time intervals through a massive amount of online interaction to reach the optimum policy.

Second, RL algorithms are not scalable to multiple problems and dynamic environments. For example, optimizing an RL algorithm in a network with a specific number of devices can not be utilized in another network with a different number of devices~\cite{da2024distributed}. Similarly, changing the characteristics of the environment, such as channel model, number of access points, and environment dimension, requires retraining the RL model from scratch, which wastes time and resources. Therefore, applying current online RL algorithms in real-world wireless systems is inefficient.

Third, most RL algorithms are not resilient to unpredictable conditions, where a minor environmental change requires retraining the RL model. In the wireless domain, resilience stands for the ability of the designed system to adapt to disturbance and maintain functionality quickly and autonomously~\cite{10187713}. Examples of unpredictable conditions include long delay intervals, sudden poor communication links outages, and unpredictable weather conditions. Hence, a resilient system includes malfunctioning detection, design parameters reformation, and normal state recovery. Therefore, resilient RL algorithms are crucial in real-time, critical applications such as UAV networks.

To address these challenges, we propose a novel offline RL algorithm that can be trained offline and requires no online interaction with the environment. We also enhance the algorithm's resilience using meta-learning, which utilizes learning across similar tasks to improve the algorithm's convergence rather than retraining each task individually.
To evaluate the performance of the proposed model, we test the algorithm on an essential application in future 6G networks, \emph{i.e.}, UAV networks.
In particular, we focus on a use-case from smart agriculture and environmental monitoring, which have emerged as critical focus areas in addressing global challenges such as food security, climate change, and sustainable resource management \cite{10680916, 9316211,RAJ2024100300}. Precision agriculture uses advanced technologies to optimize crop yield, water usage, and soil health. At the same time, environmental monitoring facilitates the adaptive management of natural ecosystems through real-time tracking of key ecological parameters. These domains demand efficient, adaptable, scalable, and resilient solutions to manage dynamic and often resource-constrained environments. UAVs, equipped with intelligent decision-making algorithms, are becoming indispensable tools in these domains due to their ability to cover vast areas efficiently and access remote or hazardous regions. 

\subsection{Offline RL and Meta-Learning}

\emph{Offline RL}~\cite{levine2020offline} is a family of RL that was proposed to overcome the problems of online RL in real-world applications. It suggests using an offline static dataset collected previously using a known behavioral policy to find the optimum policy. However, deploying existing RL and deep RL algorithms offline using static datasets without online interaction with the environment often fails to converge. This happens due to a distributional shift between the existing actions in the offline dataset and the learned actions. This problem is known as \emph{out-of-distribution (OOD)} problem and leads to overestimating the learned policies. This problem is solved in online RL by selecting the OOD actions and correcting their overestimation.

To solve the distributional shift problems in offline RL, the authors in~\cite{kumar2020conservative} propose conservative Q-learning (CQL). CQL builds over existing RL algorithms by adding a regularization parameter to the conventional optimization update to bound the influence of the OOD actions in the optimized policies. In contrast, this parameter does not affect in-distribution actions. In addition to its simple implementation over existing deep RL frameworks, such as deep Q-networks (DQNs), the CQL algorithm shows promising performance converging to the optimum policy without requiring any online environment visiting~\cite{10829755}.

Apart from offline RL, current wireless systems suffer from scalability problems. \emph{Meta-learning} is a family of learning algorithms that enable learning adaptability over adaptive and changing tasks. The most famous meta-learning algorithm is model-agnostic meta-learning (MAML)~\cite{finn2017modelagnostic}, which utilizes learning across multiple tasks to find the initialization parameters (initial neural network weights) that enable fast adaptation to new tasks through a few training iterations. This overcomes the problem of retraining the model from scratch whenever the environment changes or unpredictable conditions occur. In addition, few-shot MAML corresponds to performing fast adaptation on new tasks using a few shots of training data. This is useful when a limited amount of data is available for a task. MAML has been widely addressed in the wireless domain, such as in channel estimation, channel coding, symbol modulation and demodulation.

\subsection{Related Work}

Many works in the literature have recently adopted RL and deep RL in the wireless domain, specifically in UAV networks. Among these, the authors in~\cite{9701330} optimize the UAV optimal path to maximize the data collected from IoT devices. In~\cite{eldeeb2022multi}, the authors propose a DQN algorithm that can jointly minimize AoI and transmission power of limited-power devices. The authors in~\cite{10086052} compare centralized and decentralized deep RL techniques using soft actor-critic (SAC) for trajectory planning in integrated sensing and communications UAV network, where the work in~\cite{10109153} formulates a multi-objective deep RL algorithm for trajectory planning and beamforming design.

Apart from RL and deep RL, meta-learning has been fundamental in ensuring scalability in recent wireless applications. Many works have leveraged meta-learning techniques in the wireless domain. For example, the work in~\cite{9053252} was among the first to investigate meta-learning approaches for wireless communication. It proposes a MAML algorithm for fast training an autoencoder designed for transmitting and receiving data over fading channels. The authors in~\cite{9257198} design fast downlink beamformers using transfer learning and meta-learning. The authors in~\cite{etiabi2024metagraphlocgraphbasedmetalearningscheme} exploit meta-learning techniques with graph neural networks (GNNs) for fast wireless indoor localization. In~\cite{9495238}, deep RL is combined with meta-learning to enhance the fast adaptability of the optimizing the allocation policy a dynamic V2X network, where~\cite{9457160} proposes a multi-agent meta-RL algorithm for trajectories design of multiple UAVs. The authors in~\cite{sarathchandra2025agepowerminimizationmetadeep} jointly minimize IoT devices' AoI and transmission power using a meta-RL algorithm that adapts quickly to environments with adaptive objectives.

Although most existing RL-related works rely on online RL, offline RL has begun to get further attention in the wireless domain. The work in~\cite{eldeeb2024conser} was the first to introduce offline to the wireless domain. The authors formulate a distributional and offline multi-agent RL algorithm for planning the trajectories of multiple UAVs. The authors in~\cite{10529190} evaluate various offline RL techniques with a mixture of datasets collected from different behavioral policies for the RRM problem. In contrast, the authors in~\cite{eldeeb2024offlinedistributionalreinforcementlearning} combine distributional RL with offline RL for the RRM problem, where the work in~\cite{eldeeb2025offlinemultiagentreinforcementlearning} solves the RRM problem using multi-agent offline RL to minimize the combination of sum and tail rates jointly.

 \subsection{Main Contributions}

This work proposes a novel meta-offline MARL framework tailored for adaptive and resilient decision-making in dynamic wireless environments. The main contributions of this paper are summarized as follows.
\begin{itemize}
    \item We consider the problem of optimizing the trajectory and the scheduling policy of a UAV serving limited power sensor nodes. We formulate the problem as a joint optimization problem to minimize the AoI and the transmission power. Different tasks are defined using the trade-off between AoI and transmission power.

    \item We develop a meta-offline RL framework that integrates CQL with MAML to enhance sample efficiency and generalization, enabling rapid adaptation to new environments with limited training data.

    \item Using the MAML algorithm, we find the set of optimum initial parameters used for the CQL algorithm to be trained using a few shots of offline data points using a few stochastic gradient descent (SGD) steps.

    \item Our proposed framework's training is \textit{resilience-aware} enabling the agent to adapt to unpredictable network disruptions. Our framework ensures robust performance even in environments with adverse conditions, such as harsh weather-related link failures.
    
    \item Simulation results show that the proposed model outperforms traditional deep RL approaches regarding the resulting reward function. The proposed algorithm converges faster than the CQL algorithm with random weight initialization. In addition, it achieves the minimum possible AoI and transmission power combined compared to baseline models.
\end{itemize}
%
This is the first work to combine meta-learning with offline RL for the wireless communication domain. The rest of the paper is organized as follows. Section~\ref{sec:system_model} formulates the problem model. Section~\ref{sec:backg} describes preliminaries, whereas Section~\ref{sec:ODRL} proposes the meta-offline RL model. Simulation results are elucidated in Section~\ref{sec:results} and Section~\ref{sec:conclusions} concludes the article.

\begin{figure}[t!]
    \centering
    \includegraphics[width=1\columnwidth,trim={0cm 11cm 0cm 11cm},clip]{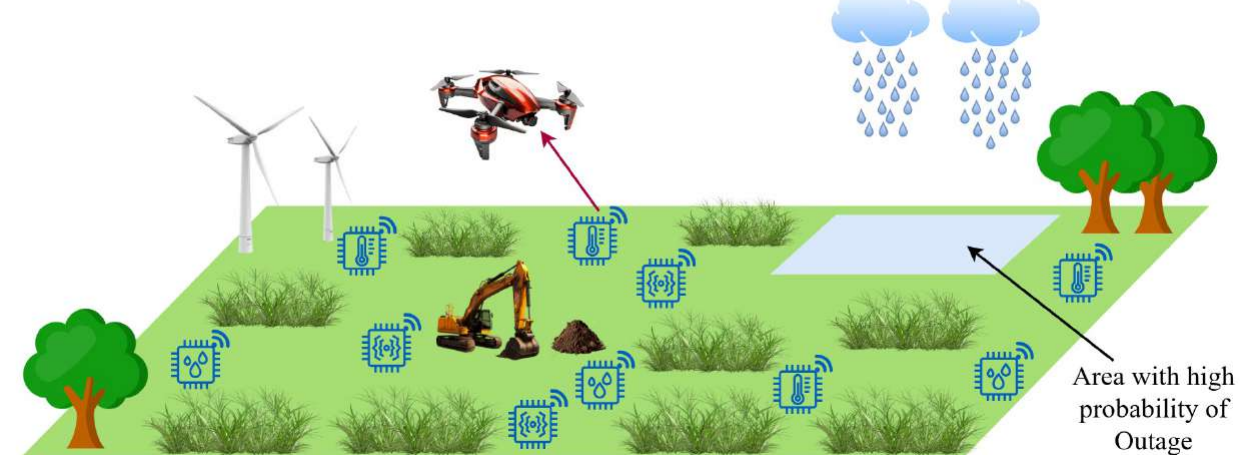} \vspace{2mm}
    \caption{Illustration of the system model. We consider smart agriculture, where a flying UAV collects information from ground nodes. In addition, sudden heavy rain occurs, which affects communication. The objective is to jointly minimize the AoI and transmission power while considering dynamic and unpredictable sources in the environment. } 
    \vspace{0mm}
    \label{UAV_Fig}
\end{figure}

\section{System Model}\label{sec:system_model}

Real-world applications in smart agriculture and environmental monitoring inspire the system model. Thus, consider the wireless network shown in Fig.~\ref{UAV_Fig} and assume a set $\mathcal{K}=\{1,2,\cdots, K\}$ of $K$ limited-power, randomly deployed, IoT devices that monitor the agricultural/environmental process(es).
We divide the network into $L \times L$ cells, where each device is positioned in the center of a cell whose length is $r$, with a coordinate $l_k = (x_k,y_k)$. The devices uplink their data to a fixed-velocity rotary-wing UAV flying with a velocity $U$ at height $h$. The position of the UAV is projected to the $2$D plane, whose coordinates, at time step $t$, are $l_u(t) = (x_u(t),y_u(t))$. At each time step $t$, the UAV serves one of the devices, where $w(t) = k$ indicates that the UAV chooses to serve device $k$. In addition, the UAV updates its current position using the movement vector $v(t)$
\begin{equation} \label{eqn:directions}
	l_u(t+1)=
	\begin{cases}
		l_u(t)+(0,r), & \quad v(t)=(0,1), \\
		l_u(t)-(0,r), & \quad v(t)=(0,-1), \\
		l_u(t)+(r,0), & \quad v(t)=(1,0), \\
		l_u(t)-(r,0), & \quad v(t)=(-1,0), \\
            l_u(t), & \quad v(t)=(0,0). \\
	\end{cases}
\end{equation}

We assume a LoS between the UAV and the devices~\cite{deep_china}. Thus, the channel gain between device $k$ and the UAV is
\begin{equation}
\label{ch_gain_d}
    g_{k,u}(t) = \frac{g_0}{h^2+||l_k-l_u(t)||^2},
\end{equation}
where $g_0$ is the channel gain at $1$ m reference distance and $||l_k-l_u(t)||^2$ is the euclidean distance between device $k$ and the UAV~\cite{eldeeb2023traffic}. Then, the transmit power of device $k$ at time $t$ is formulated using the LoS channel gain as
\begin{align}
\label{Tx_P_1}
P_k(t) \!&=\! \frac{\left(2^{\frac{M}{B}}\!-\!1\!\right)\! \sigma^2}{g_{k,u}(t)}
    \!=\! 
    \!\left(\!2^{\frac{M}{B}}\!-\!1\!\right)\!\frac{\sigma^2}{g_0}\left( h^2\!+\!||l_k\!-\!l_u(t)||^2\right),
\end{align}
where $M$ is the size of the transmitted data, $B$ is the bandwidth, and $\sigma^2$ is the noise power. Hence, the transmission power is directly related to the position of the UAV.

When the UAV enters the heavy rain area, it is affected by an attenuation factor which influences the path loss~\cite{budalal2023path,10333847}. We follow the heavy rain attenuation model described in~\cite{series2005specific}, which calculates the attenuation due to rain as 
$\gamma_R = \phi \: R^{\varphi}$,
where $R$ is the rainfall intensity set as $12.5$ mm/h. In addition, $\phi$ and $\varphi$ are fitting parameters set as in~\cite{9348592}.

The AoI quantifies how fresh the transmitted information is. It is measured as the time difference between the arrival time of a packet at a destination and its generation time at the source. Then, the AoI of device $k$ at time $t$ is updated as
\begin{equation}
\label{AoI_update}
	A_k(t) \!=\!
	\begin{cases}
		1, & \quad \text{if} \: \: w(t) = k, \\
		\text{min}\{A_{\rm max},A_k(t-1) + 1\}, & \quad \text{otherwise};
	\end{cases}
\end{equation}
where $A_{\rm max}$ is the maximum AoI in the system set to bound the complexity of the network. Hence, the AoI vector of all devices at time $t$ is $A(t) = [A_1(t), A_2(t), \cdots, A_K(t)]$.

\subsection{Problem Definition}

Often, precision agriculture applications require UAVs to perform tasks such as surveying large and/or remote areas under dynamic environmental conditions. RL-enabled UAVs can optimize their trajectories and data collection strategies to achieve application-specific goals.

Therefore, our main objective is to find the optimum policy to jointly optimize the freshness of the data collected and energy consumption. In other words, we aim to minimize the devices' weighted-sum AoI and transmission power by optimizing the UAV movement and its scheduling policy. In addition, to perform the optimization, we assume the availability of \emph{an offline dataset} $\mathcal{D}$ with few data points collected from other networks without any online interaction with the optimized network. Hence, the optimization problem is formulated as
\begin{subequations}\label{P1}
	\begin{alignat}{2}
	\mathbf{P1:}\qquad &\underset{v(t),w(t)}{\min}       &\ \ \ & \frac{1}{T}\sum_{t=1}^T\sum_{k = 1}^{K}\delta_k A_k(t) + \frac{\lambda}{K} \sum_{k = 1}^{K} P_k(t),\label{P1:a}
	\ \\
	&\text{s.t.}   &      & x_u(t), y_u(t) < L, \label{P1:b}\\
		& & & |\mathcal{D}| \leq D_{\text{const}}, \label{P1:c}
	\end{alignat}
\end{subequations}
where $\delta_k$ is a device-related importance weight and $\lambda$ is a user-chosen parameter that controls the trade-off between the AoI and the transmission power. Small $\lambda$ values indicate we minimize the AoI over the transmission power. Choosing $\lambda = 0$ eliminates the power component from the optimization problem. In contrast, large $\lambda$ values prioritize the transmission power over the AoI. Asymptotically, $\lambda \rightarrow \infty$ eliminates the AoI component from the optimization problem. The constraint~\eqref{P1:b} ensures that the UAV does not fly outside the network borders. The constraint~\eqref{P1:c} bounds the size of the offline dataset $|\mathcal{D}|$ to be greater than or equal to a certain threshold size $D_{\text{const}}$.

The optimization problem in~\eqref{P1:a} is a non-linear optimization problem, which can be solved using deep RL frameworks, such as DQNs. However, relying only on offline static data points without online interaction poses serious challenges. In this work, we overcome these challenges by combining offline RL with meta-learning. Offline RL solves the optimization problem using historical datasets collected from prior UAV operations, eliminating the need for costly and risky online training. Meanwhile, meta-RL ensures that the system can quickly adapt to new objectives by utilizing similar networks to optimize the use of a few (online) data points. In addition, meta-learning ensures the resilience of the RL algorithm to adapt to unpredictable conditions.

\section{Background}\label{sec:backg} 
This section introduces the basics of deep RL, offline RL, and meta-learning. These preliminaries will help present the proposed meta-offline RL algorithm in the next section.

\begin{algorithm}[!t]
\SetAlgoLined

\textbf{Define} the hyperparameters $\lambda$, $\epsilon$, $\gamma$, $\eta_{\text{QL}}$, number of episodes $I$, and length of each episode $T$.

Initialize the Q-network and target Q-network with weights $\mathrm{w^0}$.

Initialize the experience replay buffer.

\For{\text{episodes} $i$ in $\{1, \cdots,I\}$}{
    \For{\text{time} $t$ in $\{1, \cdots,T\}$}{
        Explore a random action $a^{\prime}$ with probability $\epsilon$ or select optimal action $a = \max_a Q(s^{\prime},a^{\prime})$ with probability $1-\epsilon$.
        
        Save $\langle s,a,r,s^{\prime} \rangle$ in the replay buffer.
        
        Sample a mini-batch from the buffer.
    
        Compute the loss in~\eqref{bellman_error}.
        
        Update the weights of the Q-network and the target Q-network.
    }
}

\textbf{Return} Converged Q-network (Q-function $Q(s,a)$) with optimal weights $\mathrm{w}^*$

\caption{Deep Q-network (DQN) algorithm.}
\label{alg1}  \vspace{0mm}
\end{algorithm}

\subsection{Deep Reinforcement Learning}
In the previous section, we formulated the policy optimization problem of minimizing the IoT devices' average AoI and transmission power. This problem can be viewed as a Markov decision process (MDP). Generally, MDPs compose of the tuple $\langle s(t), a(t), r(t), s(t+1) \rangle$, where $s(t)$ is the current state, $a(t)$ is the selected action, $r(t)$ is the immediate reward, and $s(t+1)$ is the next state resulted from taking action $a(t)$ at state $s(t)$. In RL, the goal is to find the optimum policy $\pi^*$ that maximizes the accumulative rewards.

In our UAV problem, the MDP can be elucidated as follows:
\begin{itemize}
    \item \textbf{States:} At time instant $t$, the state space consists of the UAV location $l_u(t)$ and the individual AoI of each device $A(t)$. Hence, the detailed state space is $s(t) = [x_u(t), y_u(t), A_1(t), A_2(t), \cdots, A_K(t)]$, whose length is $(2+K)$.

    \item \textbf{Actions:} At time instant $t$, the action space consists of the movement direction of the UAV $v(t)$ and the chosen device to serve $w(t)$. The detailed action space is $a(t) = [v(t), w(t)]$, where the dimension of all possible available actions is $(5 \times K)$.

    \item \textbf{Rewards:} The reward function is formulated in a way to serve the optimization problem in~\eqref{P1}
    \begin{equation}
        r(t) = -\frac{1}{T}\sum_{t=1}^T\sum_{k = 1}^{K}\delta_k A_k(t) + \frac{\lambda}{K} \sum_{k = 1}^{K} P_k(t),
    \end{equation}
    where the negative sign in the reward function ensures jointly minimizing the AoI and the transmission power by maximizing the set of received rewards.    
\end{itemize}

Q-learning is a famous RL algorithm that efficiently solves MDPs iteratively. It utilizes the state-action value function (Q-function) $Q(s, a)$, which is an evaluation function that evaluates the set of available actions at each state by computing the expected accumulative rewards (return). The Q-learning iteratively finds the optimum Q-function $Q^*(s,a)$, which corresponds to the optimum policy, as follows
\begin{equation}
Q\left(s,a\right) \!=\!   Q\left(s,a\right) \!+\! \eta_{\text{QL}}  \!\left(r \!+\!  \gamma \: \max_{a^{\prime}} Q\left(s^{\prime}, a^{\prime}\!\right) \!-\! Q\left(s, a\right)\!\right)\!,
\end{equation}
where $\eta_{\text{QL}}$ is a chosen learning rate, $s^{\prime}$ is the next state, $a^{\prime}$ is the next action, $\gamma \: Q\left(s^{\prime},a^{\prime}\right)$ is the discounted future state-action value function and $\gamma$ is the discount factor, which controls the weights of future rewards compared to immediate rewards in the Q-learning solution.

Despite its MDP efficiency, it suffers from large-dimensional problems like the one we solve in this work. DQNs combine Q-learning with efficient DNNs, which act as function approximators to the Q-function $Q(s, a)$ and the target Q-function $\hat{Q}(s^{\prime},a^{\prime})$. It proposes storing previous experiences $\langle s, a,r,s^{\prime} \rangle$ in an experience replay buffer to estimate the Q-function better. At each step, it samples a mini-batch from the experience replay to update the Q-function and the target Q-function using the loss
\begin{align}
\label{bellman_error}
    \mathcal{L}^{\text{DQN}}_{\mathrm{w}} (Q, \hat{Q}) = \:& \hat{\mathbb{E}} \left[ \left(  r +\gamma \max_{a^{\prime}} \hat{Q}(s^{\prime},a^{\prime}) 
   - Q(s,a) \right)^2 \right],
\end{align}
where $\mathrm{w}$ is the set of weights of the Q-network, $\hat{\mathbb{E}}[\cdot]$ is the empirical average over the sampled experience $(s,a,r,s')$, $\hat{Q}(s,a)$ and $\hat{Q}(s^{\prime},a^{\prime})$ are modeled using neural networks, and $r +\gamma \max_{a^{\prime}} \hat{Q}(s^{\prime},a^{\prime}) - Q(s,a) $ is known as the TD-error. In typical RL problems, an exploration rate $\epsilon$ is defined, which decays with time to ensure enough exploration of the environment. With a probability $\epsilon$ the agent explores a random action, while it selects the action that maximizes the Q-function with a probability $1-\epsilon$. Algorithm~\ref{alg1} summarizes the DQN algorithm.

\begin{figure}[t!]
    \centering
    \includegraphics[width=1\columnwidth,trim={0 5cm 0 2.5cm},clip]{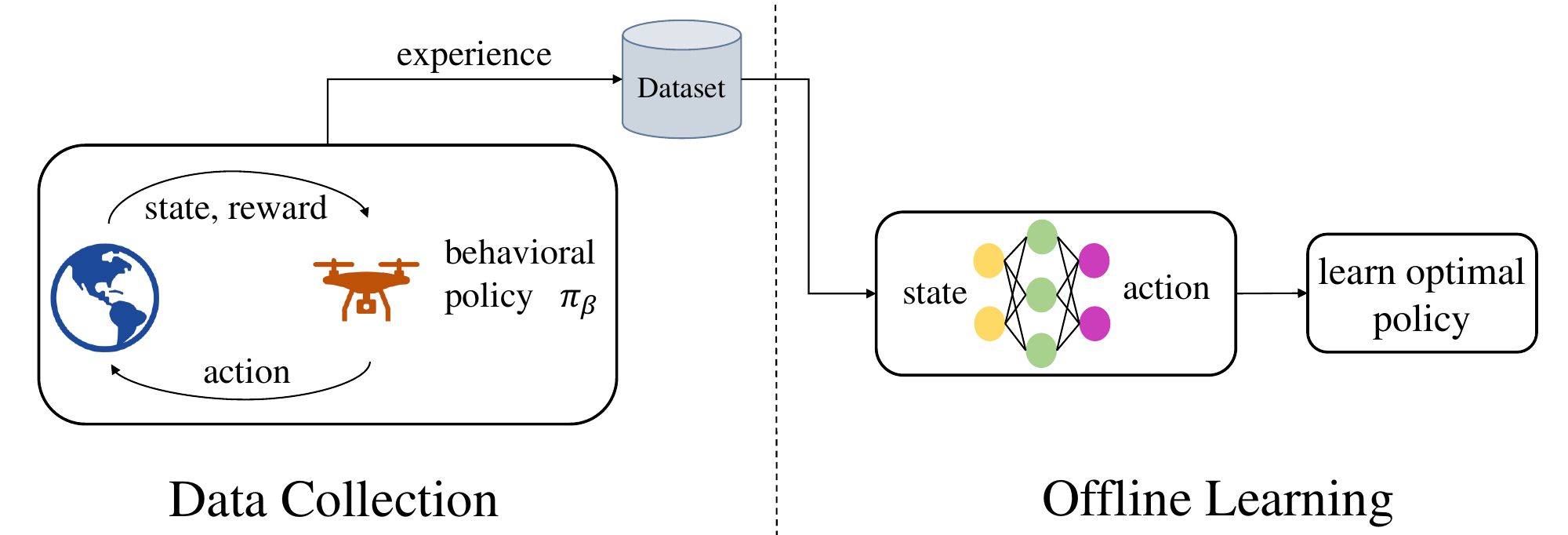} 
    \caption{Illustration of Offline RL, which involves two phases: data collection and offline learning. In the data collection phase, fixed datasets are collected using behavioral policies. A learning model uses a static offline dataset to find the optimum policy in the offline learning phase.} 
    \vspace{0mm}
    \label{Offline_RL_Fig}
\end{figure}

\subsection{Offline Reinforcement Learning}
The aforementioned DQN algorithm is an off-policy online DRL algorithm as it relies heavily on online interaction with the environment besides sampling offline data from the replay buffer. However, online interaction with the environment might not be feasible due to the expensive online data collection, nor safe due to the uncertainties in the environment. Offline RL solves this problem by suggesting collecting offline datasets from previous behavioral policies $\pi_{\beta}$ (or even a random policy) to be used to find the optimum policy without any interaction with the environment~\cite{dabney2018implicit}.

However, deploying the presented DQN algorithm offline with a static dataset poses many problems. First, the lack of online interaction introduces out-of-distribution (OOD) actions, which come from the differences between the available and learned policies. Second, offline RL algorithms usually overestimate the quality of the optimized actions due to the limited amount of data. These problems are solved in off-policy online DRL algorithms (DQNs) by exploring experiences online. Therefore, deploying the DQN algorithm using only offline datasets generally fails.

\begin{algorithm}[!t]
\SetAlgoLined

\textbf{Define} the hyperparameters $\lambda$, $\epsilon$, $\gamma$, $\eta_{\text{QL}}$, $\alpha$ and number of training epochs $E$.

Initialize the Q-network with weights $\mathrm{w^0}$.

Collect an offline dataset $\mathcal{D}$ using $\pi_{\beta}$.

\For{\text{epoch} $e$ in $\{1, \cdots,E\}$}{

Sample a batch $\mathcal{B}$ from the dataset $\mathcal{D}$.

Estimate the CQL loss $\mathcal{L}_{\text{CQL}}$ using~\eqref{CQL_loss_eq}.

Update the weights of the Q-network.

}

\textbf{Return} Converged Q-network (Q-function $Q(s,a)$) with optimal weights $\mathrm{w}^*$.

\caption{Conservative Q-learning (CQL) algorithm for Offline RL.}
\label{CQL_alg}  \vspace{0mm}
\end{algorithm}

To this end, \emph{Conservative Q-learning (CQL)}~\cite{kumar2020conservative} is an offline RL algorithm that adjusts the traditional Q-learning algorithm for offline training. It overcomes the distributional shift problem by adding a regularization term to the bellman update. The regularization term penalizes the OOD distribution, forcing the selected actions to be as close as possible to the offline data set $\mathcal{D}$ collected using a behavioral policy $\pi_{\beta}$. The CQL loss is formulated as
\begin{align}
 \label{CQL_loss_eq}
\mathcal{L}^{\text{CQL}}_{\mathrm{w}} (Q, \hat{Q})=&\: \frac{1}{2}\mathcal{L}^{\text{DQN}}_{\mathrm{w}} (Q, \hat{Q})\\ &+\alpha \hat{\mathbb{E}} \bigg[ \log \sum_{\tilde{a}}
\exp \bigl( Q(s,\tilde{a}) \bigr) 
    - \ Q(s,a)  \bigg], \nonumber
\end{align}
where $\alpha > 0$ is a user-chosen parameter called the conservative parameter and $\tilde{a}$ ensures that all possible actions are evaluated. \textbf{Algorithm~\ref{CQL_alg}} illustrates the CQL algorithm.

\subsection{Meta-Learning}
Meta-learning has been a fundamental element in many wireless communication domains in recent years due to its ability to find a scalable, transferable, and resilient solution. The most famous definition for meta-learning is learning to learn, as it utilizes learning across different tasks to enhance the convergence rate of new tasks. In our problem, for example, selecting different values for $\lambda$ in~\eqref{P1} corresponds to a different set of functions as $\lambda$ controls the trade-off between AoI and transmission power. Hence, each time we change $\lambda$, we change the objective of the problem. To this end, meta-learning can learn from environments with different $\lambda$ values to speed up the learning convergence of the UAV when selecting a new value for $\lambda$~\cite{eldeeb2024semanticmetasplitlearningtinyml}. Another example is a sudden environmental condition, such as areas with poor communication links due to weather conditions. Meta-learning can ensure resilient performance by transferring learning across different tasks to adapt to new unpredictable sources quickly.

\begin{algorithm}[!t]
\SetAlgoLined

\textbf{Define} the hyperparameters $\eta_{\text{inner}}$, $\eta_{\text{outer}}$, $T$, $p(\tau)$ and number of meta training epochs $E_{\text{meta}}$.

Initialize the model weights $\mathrm{w^0}$.

\For{\text{epochs} $e$ in $\{1, \cdots,E\}$}{


\For{\text{task} in $\{\tau_1, \cdots,\tau_T\}$}{

Sample $k$ shots from the support set.

Update the model weights $\mathrm{w}$ using the support set and the update in~\eqref{task_update}.
}
    
Calculate the meta-loss $\mathcal{L}_{\text{meta}}$ using~\eqref{meta_loss}.

Update the initial weights $\mathrm{w}^0$ using~\eqref{meta_optimization}.

}

\textbf{Return} model converged initial weights $\mathrm{w}^0$.

\caption{Model agnostic meta-learning (MAML) algorithm~\cite{finn2017modelagnostic}.}
\label{Meta_Alg}  
\end{algorithm}

\emph{Model agnostic meta-learning (MAML)}~\cite{finn2017modelagnostic} is a well-known meta-learning algorithm that utilizes learning across different tasks to find the set of initial weights that can lead to a faster convergence on new unseen tasks. Consider the set of $T$ unique tasks $\{ \tau_1, \cdots, \tau_T\}$ sampled randomly and independently from a task distribution $p(\tau)$. Let the set of initial weights of a model be $\mathrm{w}^0$. During the meta-training phase, the available dataset is divided into \textit{support and query sets}. The \textit{support set} is used for training each task, where a few \emph{shots} (data points) of length $K$ ($K$-shots) are sampled for training. The set of initial weights is trained over each sampled task as follows
\begin{equation}
    \label{task_update}
    \mathrm{w} \leftarrow \mathrm{w} - \eta_{\text{inner}}\nabla_{\mathrm{w}} \mathcal{L}_{\mathrm{w}} (\tau_i),
\end{equation}
where $\eta_{\text{inner}}$ is the inner meta-learning rate and $\mathcal{L}_{\mathrm{w}} (\tau_i)$ is an appropriate loss function designed for task $\tau_i$. After training each task, we estimate the individual losses using \textit{the query set} as follows
\begin{equation}
    \label{meta_loss}
    \mathcal{L}_{\text{meta}} = \sum_{i=1}^{T} \mathcal{L}_{\mathrm{w}} (\tau_i),
\end{equation}
where $\mathcal{L}_{\text{meta}}$ is the meta-loss. Afterward, a global update is performed over the initial weights using the meta-loss
\begin{equation}
    \label{meta_optimization}
    \mathrm{w}^0 \leftarrow \mathrm{w}^0 - \eta_{\text{outer}}  \nabla_{\mathrm{w^0}} \mathcal{L}_{\text{meta}},
\end{equation}
where $\eta_{\text{outer}}$ is the outer meta-learning rate. \textbf{Algorithm~\ref{Meta_Alg}} illustrates the MAML algorithm.

After convergence, a new unseen task is sampled for the meta-testing phase, in which the available dataset is also divided into a support set and a query set. The set of optimal initial weights $\mathrm{w}^0$ is trained through a few SGD steps using $k$ shots sampled from the support set. Then, the query set is used for testing. 

\begin{algorithm}[!t]
\SetAlgoLined

\textbf{Define} the hyperparameters $\epsilon$, $\gamma$, $\eta_{\text{QL}}$, $\eta_{\text{inner}}$, $\eta_{\text{outer}}$, $\alpha$, $T$, task distribution $p(\tau)$ with unique $\lambda$ values, number of training epochs $E$ and number of meta training epochs $E_{\text{meta}}$.

Initialize the Q-network with weights $\mathrm{w^0}$.

Collect an offline dataset $\mathcal{D}$ using $\pi_{\beta}$ and divide it into support and query sets.

\For{\text{epochs} $e$ in $\{1, \cdots,E_{\text{meta}}\}$}{


\For{\text{task} in $\{\tau_1, \cdots,\tau_T\}$}{

Sample $k$ shots from the support set.

Update the model weights $\mathrm{w}$ using the support set and the update in~\eqref{task_update_prop}.
}
    
Calculate the meta-loss $\mathcal{L}_{\text{meta}}$ using~\eqref{meta_RL_loss}.

Update the initial weights $\mathrm{w}^0$ using~\eqref{meta_RL_optimization}.

}

\textbf{Return} model converged initial weights $\mathrm{w}^0$.

\textbf{Define} a new unseen task corresponds to a new $\lambda$ value and use the converged initial weights $\mathrm{w}^0$.

\For{\text{epoch} $e$ in $\{1, \cdots,E\}$}{

Sample $k$ shots from the support set.

Estimate the CQL loss $\mathcal{L}_{\text{CQL}}$ using~\eqref{CQL_loss_eq}.

Update the weights of the Q-network.

}

\textbf{Return} Converged Q-network (Q-function $Q(s,a)$) with optimal weights $\mathrm{w}^*$

\caption{The proposed few-shot meta-offline RL (CQL-MAML) algorithm.}
\label{Meta_off_Alg}  
\end{algorithm}

\section{Few-Shot Meta-Offline RL}\label{sec:ODRL}
\begin{figure*}[h!]
    \centering
    \includegraphics[width=1.9\columnwidth,trim={0 0cm 0 0cm},clip]{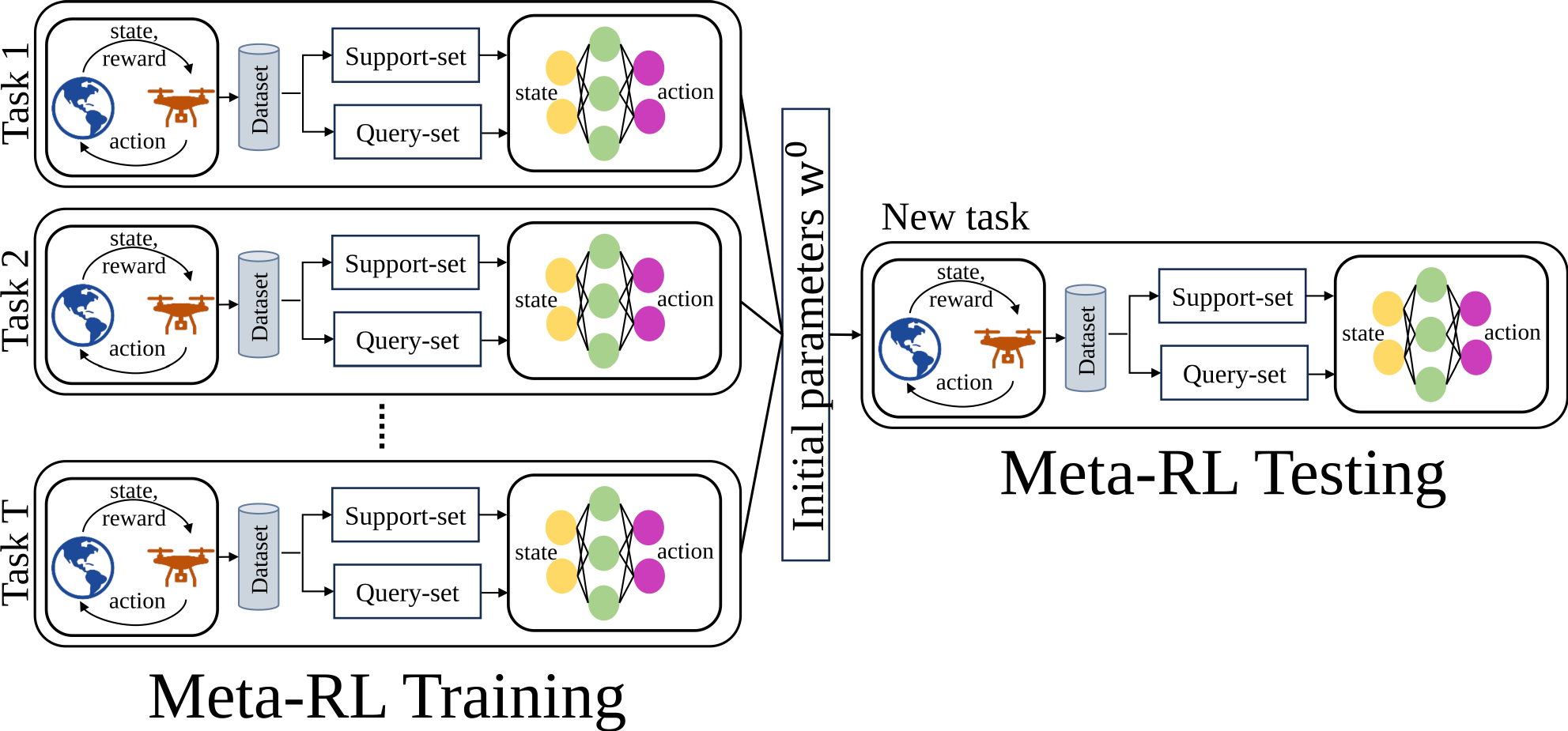} 
    \caption{Illustration of the proposed CQL-MAML algorithm, composed of \textit{meta-RL training} and \textit{testing phases}. The former utilizes offline training, using the CQL algorithm, across different tasks (environments) with different objectives to find the optimum initial parameters. In contrast, the latter performs a few offline SGD steps over the reached weights on a new unseen task.} 
    \vspace{0mm}
    \label{Offline_meta_RL_Fig}
\end{figure*}

This section presents the proposed few-shot meta-offline RL algorithm for UAV policy optimization. We combine the offline CQL algorithm with the MAML algorithm to train a UAV agent with limited offline data utilizing offline learning across similar tasks. We generate different tasks, each corresponding to a unique environment with a unique $\lambda$ value. We randomly initialize the Q-network weights $\mathrm{w^0}$ during meta-offline RL training. Then, we update these weights internally for each task using the CQL loss as follows
\begin{equation}
    \label{task_update_prop}
    \mathrm{w} \leftarrow \mathrm{w} - \eta_{\text{inner}}\nabla_{\mathrm{w}} \mathcal{L}^{\text{CQL}}_{\mathrm{w}} (Q, \hat{Q};\tau_i).
\end{equation}
Similar to~\eqref{meta_loss}
\begin{equation}
    \label{meta_RL_loss}
    \mathcal{L}_{\text{meta-RL}} = \sum_{i=1}^{T} \mathcal{L}^{\text{CQL}}_{\mathrm{w}} (Q, \hat{Q};\tau_i),
\end{equation}
which is used to update the initial weights of the Q-network
\begin{equation}
    \label{meta_RL_optimization}
    \mathrm{w}^0 \leftarrow \mathrm{w}^0 - \eta_{\text{outer}}  \nabla_{\mathrm{w^0}} \mathcal{L}_{\text{meta-RL}}.
\end{equation}

During the meta-offline RL testing phase, we use the converged initial parameters on a new sampled task with a new $\lambda$ value and perform a few SGD steps using the traditional CQL algorithm. The proposed few-shot meta-offline RL algorithm is presented in \textbf{Algorithm~\ref{Meta_off_Alg}}.

\section{Numerical Results}\label{sec:results}
In this section, we compare the performance of the proposed few-shot meta-offline RL algorithm to the baselines designed for optimizing the UAV trajectory and its scheduling policy to jointly minimize the AoI and the transmission power of limited-power devices. First, we show the implementation and key simulation metrics. Second, we evaluate the proposed algorithm's scalability by changing the problem's objective. Then, we consider the algorithm's resilience in sudden heavy rain conditions.

\subsection{Implementation}
We consider a $1000$ m $\times$ $1000$ m square area with $L = 10$ cells and $K = 10$ IoT devices. Consider an episodic scenario, where the length of an episode $T=100$ time steps. At the beginning of each episode, the UAV starts from a random position. We use $2$ hidden layers in the implemented neural network for the proposed model and baseline architectures, \emph{i.e.}, DQN, and CQL. All simulations are performed on a single NVIDIA Tesla V100 GPU using Pytorch framework~\cite{NEURIPS2019_bdbca288}. Table~\ref{UAV_Parameters} shows the parameters used in the simulation.

Environments are defined by the user-chosen $\lambda$ value that controls the trade-off between the AoI and the transmission power. Hence, each meta-task corresponds to a unique environment with a unique $\lambda$ value. For each environment, a corresponding offline dataset is collected using the replay buffer of an online DQN agent. During meta-training, some tasks (datasets) are sampled to optimize the initial weights of the Q-network to be tested on a new unseen dataset. We compare the proposed CQL-MAML model to baseline offline models, such as CQL, DQN, and DQN-MAML.
\begin{table}[t!]
\centering
\caption{Simulation parameters and neural network hyperparameters.}
\label{UAV_Parameters}
\begin{tabular}{cc|cc}
\toprule
\textbf{Parameter}                                    & \textbf{Value} & \textbf{Parameter}                                    & \textbf{Value} \\ \midrule
\midrule

$g_0$ & $30$ dB & $\alpha $ & $1$\\
$B$ & $1$ MHz & $\gamma$ & $0.99$\\
$h$ & $100$ m & $r$ & $100$ m\\
$M$ & $5$ Mb & $\sigma^2$ & $-100$ dBm\\
$\gamma$ & $0.99$ & $A_{\max}$ & 100 \\
$\eta_{\text{inner}}$ & $10^{-2}$ & $\eta_{\text{outer}}$ & $10^{-3}$ \\
Meta-epochs & $150$ & Optimizer & Adam \\

\bottomrule
\end{tabular} 
\end{table}

\begin{figure}[t!]
    \centering    \includegraphics[width=1\columnwidth,trim={0cm 0 0cm 0},clip]{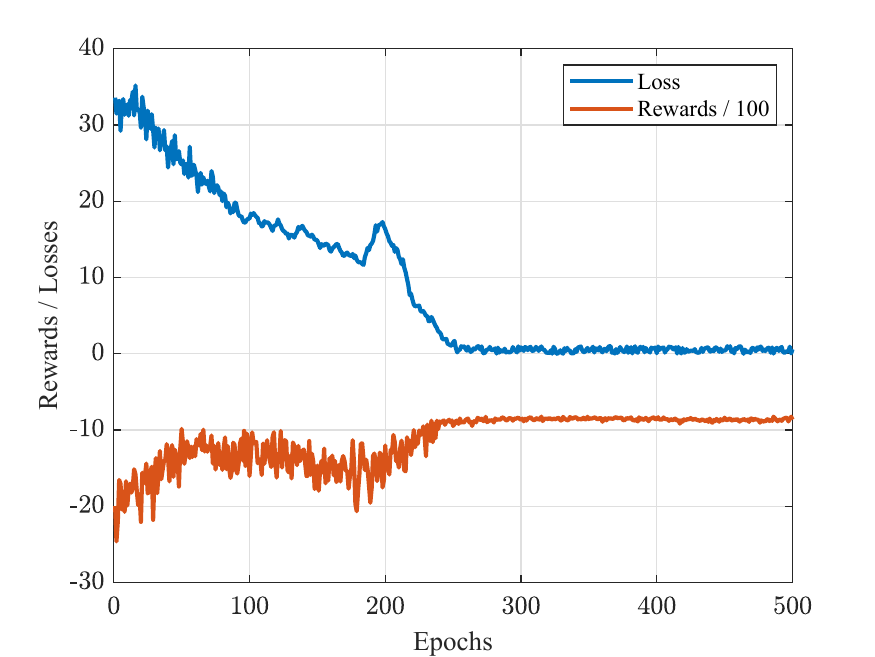} 
    \caption{An illustration of the meta-training performance of the proposed CQL-MAML algorithm using $8$ meta-tasks and an offline dataset with $500$ data points. Both loss and rewards (normalized by $100$ converge to their minimum and maximum values, respectively.}
    \vspace{0mm}
    \label{Meta_Loss_Rewards}
\end{figure}

\subsection{Adaptive Objective}

In the first experiment in Fig.~\ref{Meta_Loss_Rewards}, we show the training performance of the proposed CQL-MAML in terms of the loss convergence and achieved rewards of meta-training tasks. As shown in Fig.~\ref{Meta_Loss_Rewards}, we report the per-task loss, the mean of the losses of meta-training tasks, as a function of training epochs. We use $8$ meta-tasks to perform the training using a dataset with $500$ experiences sampled from the experience replay of an online DQN agent. The convergence of the loss demonstrates the success of the CQL-MAML in finding good initial weights that work well for all meta-training tasks. Similarly, the figures show the per-task rewards as a function of training epochs. The proposed model can find the set of initial weights that can achieve the optimal policy that maximizes the rewards across meta-training tasks.
\begin{figure}[t!]
    \centering    \includegraphics[width=1\columnwidth,trim={0cm 0 0cm 0},clip]{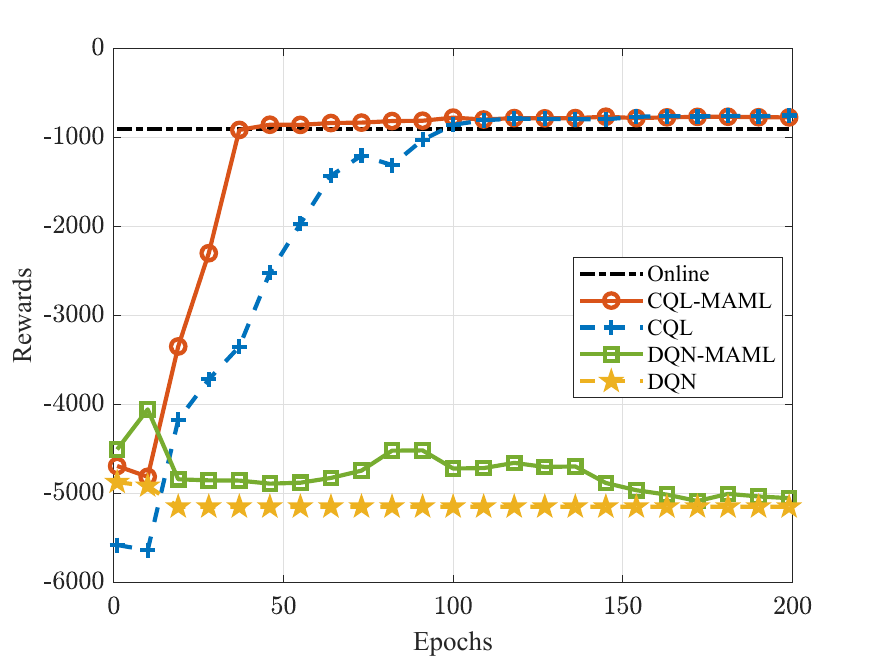} 
    \caption{The meta-testing convergence of the proposed CQL-MAML algorithm compared to baseline schemes in a new unseen task after training the initial weights using $8$ meta-tasks and an offline dataset with $500$ data points.}
    \vspace{0mm}
    \label{Converg}
\end{figure}

In the next experiment, we test the converged initial weights of the Q-network through offline SGD steps on a new unseen task (environment) with a unique objective ($\lambda$). We use an offline dataset with $500$ experiences sampled from the experience replay of an online DQN agent. As shown in Fig.~\ref{Converg}, the baseline models DQN and DQN-MAML fail entirely to converge due to the distributional shift problems. The CQL algorithm (using random Q-network initialization) needs more than $100$ training epochs to converge to the optimal policy. In contrast, the proposed CQL-MAML, trained over $8$ tasks, requires less than $40$ epochs to converge to the optimal policy. It is worth mentioning that even without any exploration, both CQL and CQL-MAML find a better policy that scores better rewards than an online DQN agent.
\begin{figure}[t!]
    \centering    \includegraphics[width=1\columnwidth,trim={0cm 0 0cm 0},clip]{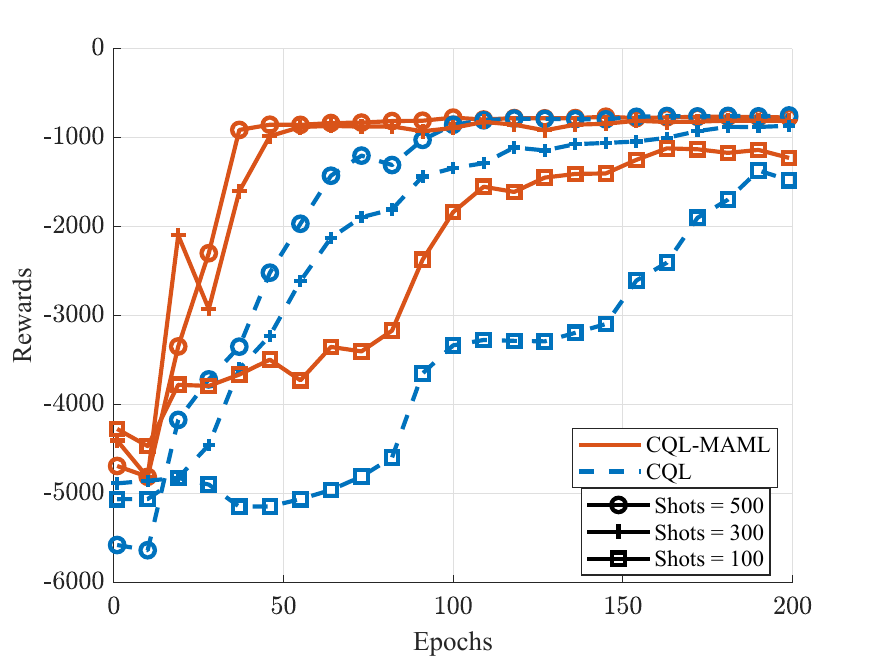} 
    \caption{A visualization of the effect of the size of the offline dataset ($\text{shots} = \{100,300,500\}$) on the proposed CQL-MAML algorithm compared to the CQL algorithm. Larger datasets enhance the convergence speed of both algorithms.}
    \vspace{0mm}
    \label{Shots}
\end{figure}

Fig.~\ref{Shots} leverages the effect of the size of the offline dataset on both CQL and the proposed CQL-MAML algorithms. 
For clarity of visualization, we omit the performance of both DQN and DQN-MAML algorithms due to their poorer performance than all other algorithms. We can notice that regardless of the size of the dataset, the proposed CQL-MAML consistently outperforms the traditional CQL algorithm in terms of the rewards achieved and the training epochs needed. This highlights the power of finding good initial weights on the convergence speed. In addition, as the number of shots (size of the dataset) increases, both CQL and CQL-MAML record better rewards.
The CQL algorithm seems to be heavily affected by the dataset size due to the large gap between the training rewards in different dataset sizes. This is not the case with CQL-MAML, which has a more relatively bounded performance.

\begin{figure}[t!]
    \centering    \includegraphics[width=1\columnwidth,trim={0cm 0 0cm 0},clip]{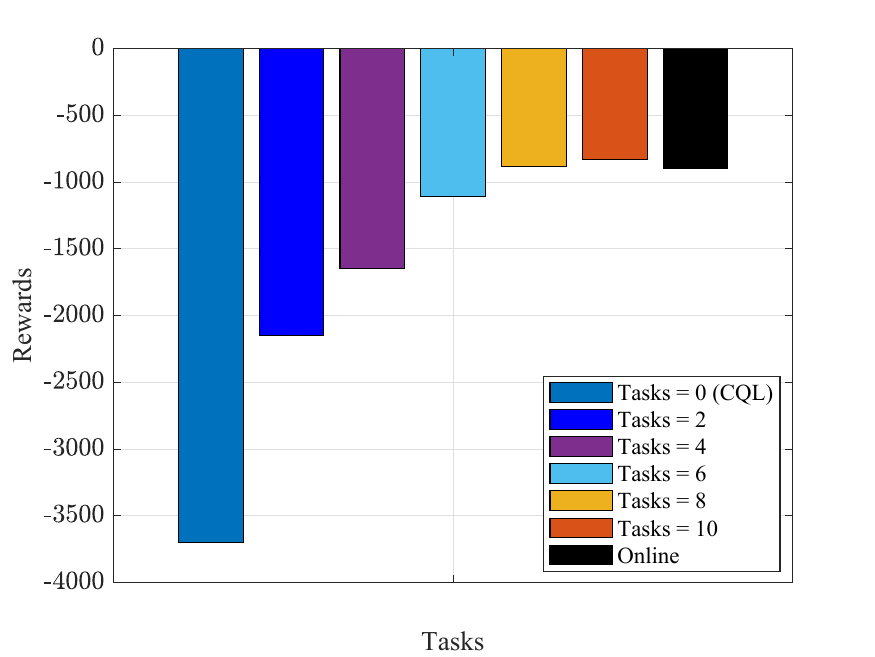} 
    \caption{A visualization of the effect of the number of tasks ($\text{tasks} = \{2,4,6,8,10\}$) on the proposed CQL-MAML algorithm compared to online and CQL rewards. A larger number of tasks enhance the convergence of CQL-MAML rewards.}
    \vspace{0mm}
    \label{Tasks}
\end{figure}

Fig.~\ref{Tasks} exploits the importance of the number of meta-training tasks on the performance of the proposed CQL-MAML. Given that regardless of the number of tasks, the CQL-MAML will eventually, after enough epochs, record the rewards of performing meta-tasks over a new unseen task with only $40$ epochs. To enable the CQL-MAML to surpass online DQN using $40$ epochs, we need to perform meta-training over at least $8$ training tasks. However, using a lower number of tasks in meta-training (as low as $2$ tasks) still results in better reward performance than the conventional CQL algorithm (which we refer to here as the $0$ tasks CQL-MAML algorithm).
\begin{figure}[t!]
    \centering    \includegraphics[width=1\columnwidth,trim={0cm 0 0cm 0},clip]{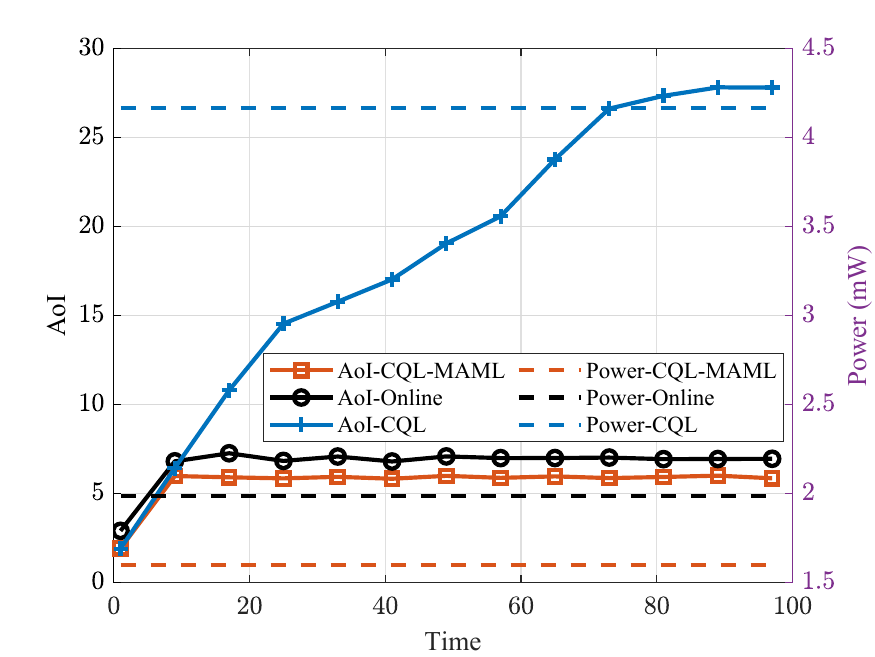} 
    \caption{The AoI and transmission power achieved after meta-testing using $50$ training epochs on a new unseen task. The proposed CQL-MAML algorithm has the least joint AoI and transmission power.}
    \vspace{0mm}
    \label{AoI_Power}
\end{figure}

In Fig.~\ref{AoI_Power}, we visualize the resulting AoI and transmission power after the convergence of the meta-testing phase over a new unseen task ($\lambda = 300$) of the proposed meta-CQL algorithm compared to CQL and online DQN. In this experiment, we set the number of meta-training tasks to $8$, the size of the used dataset to $500$, and the number of training epochs to $50$. The proposed meta-CQL algorithm achieves lower AoI ($6$) and transmission power ($1.6$ mW) compared to CQL, which scores a very high AoI ($28$) and power ($6$ mW). In addition, it has lower AoI and transmission power than online DQN, which has $7$ units of AoI and consumes $2.2$ mW of power. This highlights that the proposed offline method can achieve better policies than online RL without interacting with the environment. 

\subsection{Resilient Design}
\begin{figure}[h!]
    \centering
    \subfloat[Convergence\label{conv_out}]{\includegraphics[width=1\columnwidth]{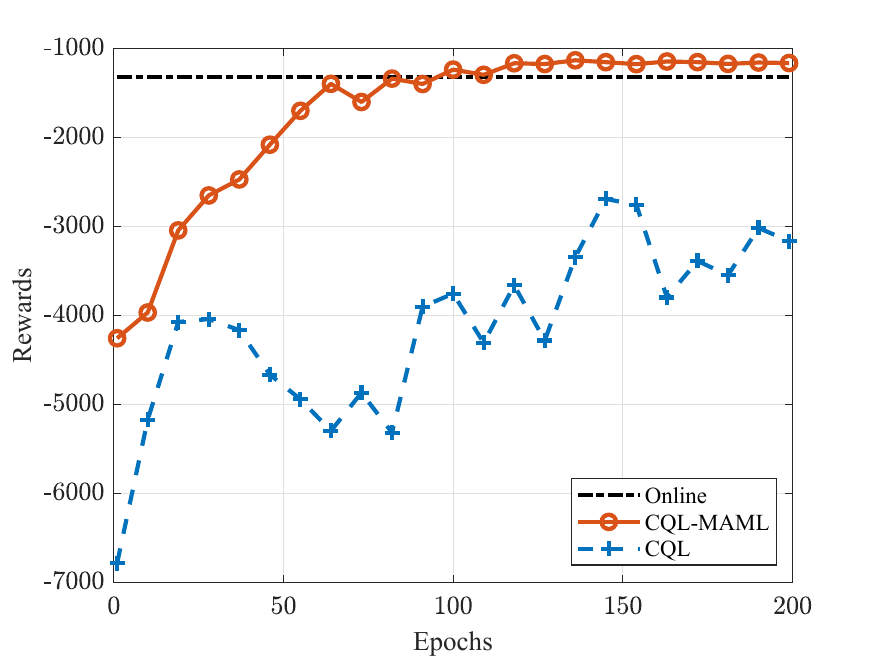}}\\
    \subfloat[Number of outages\label{Outages}]{\includegraphics[width=1\columnwidth]{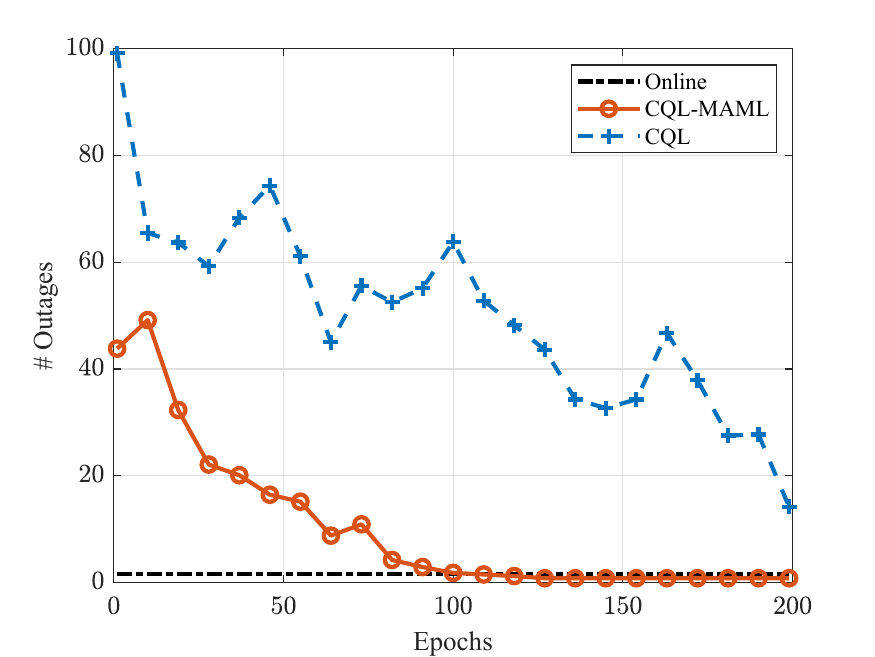}}\\
    \subfloat[Trajectories\label{Traject}]{\includegraphics[width=1\columnwidth]{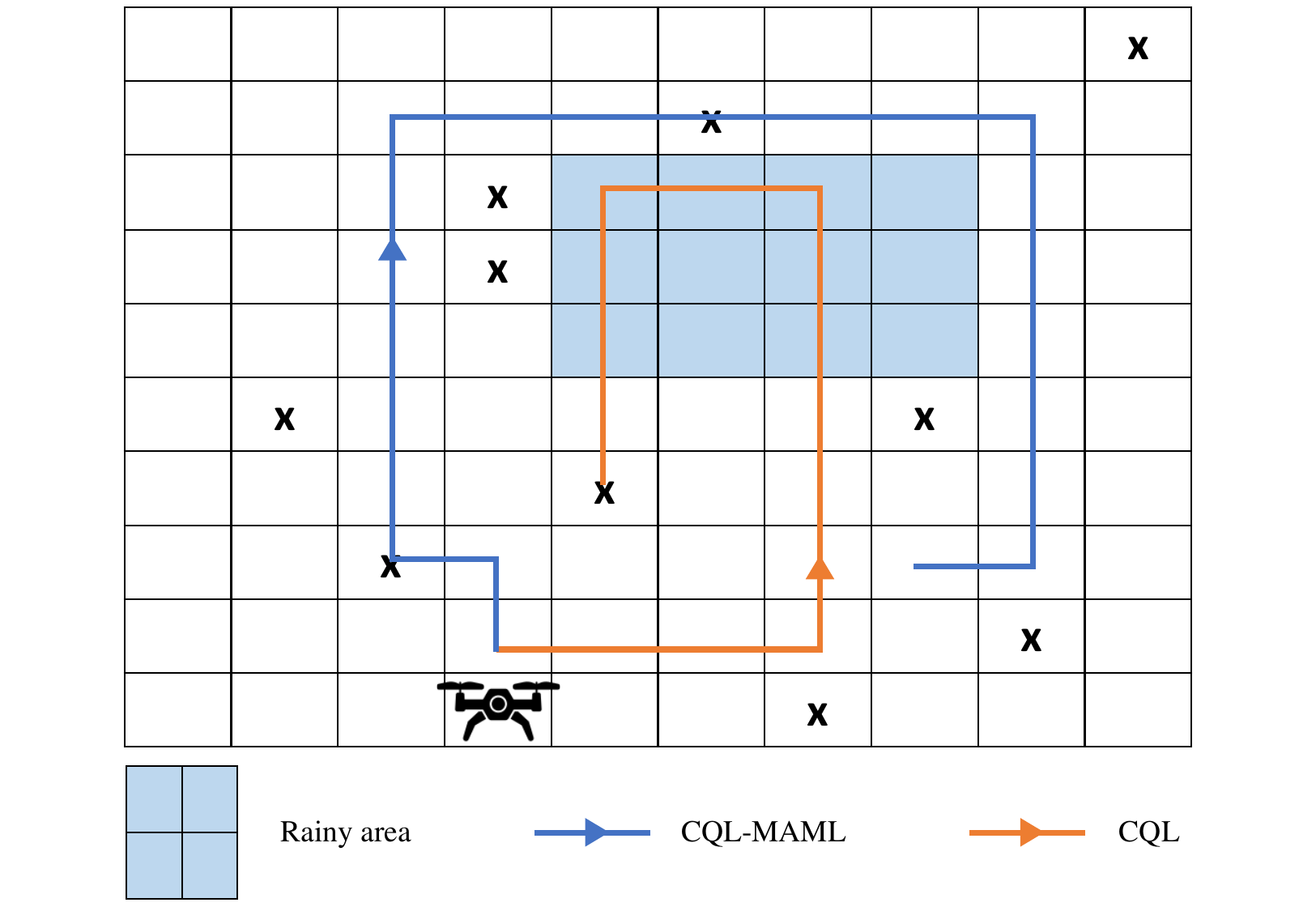}}
    \vspace{-1mm}
    \caption{Performance evaluation of the proposed algorithm in detecting outages (a) convergence, (b) number of outages, and (c) UAV trajectories.}
    \label{Conv_Out_Traj} 
\end{figure}
In this experiment, we test the resilience of the proposed algorithm to achieve a robust and reliable performance. We consider a source of outage (\emph{e.g.,} heavy rain) in the network that leads to poor links between the UAV and ground devices. The location of this source quickly changes, and the UAV needs to adapt to such unpredictable conditions. 
To validate the resilience of our approach, we conducted meta-training in outage-free environments and then evaluated the model on five unseen tasks where outages occur unpredictably. This tests the proposed model's resilience in sudden, unprecedented conditions. The performance of the proposed CQL-MAML, CQL, and online RL are shown in Fig.~\ref{Conv_Out_Traj}. Similar to the previous experiment, we exploit static datasets for training and testing, where each dataset consists of $500$ data points.
%

Fig.~\ref{conv_out} demonstrates the convergence of the learning schemes. Our model (CQL-MAML) converges to higher reward values than the online case, consuming less than $100$ epochs. Without using MAML, CQL could not reach convergence in $200$ epochs. Fig.~\ref{Outages} evaluates the number of experienced failures due to entering the source of outage area as a function of training epochs (each point is evaluated over $1000$ monte-Carlo loops over the $5$ new unseen tasks. After $100$ training epochs, the proposed CQL-MAML avoids entering the outage area and, thus, experiences almost zero outage (similar to the online case). In contrast, CQL experiences around $20$ outages from entering the area. This highlights the resilience of the proposed algorithm to recover and stabilize against unpredictable conditions. Fig.~\ref{Conv_Out_Traj} visualizes a snapshot from the trajectories of CQL-MAML and CQL. The proposed CQL-MAML avoids entering the heavy rain area and moving around its corners, while CQL impairs the communication links by spending long intervals inside it.
The proposed approach effectively avoids outage-prone areas, ensuring reliable data collection and transmission in UAV-based networks. This is crucial for precision agriculture applications and can be extended to other verticals, such as disaster response.

\section{Conclusions}\label{sec:conclusions} 

In this paper, we designed a novel and resilient few-shot meta-offline RL algorithm to plan the UAV trajectory and its scheduling policy to jointly minimize the AoI and the transmission power of ground IoT devices. In summary, We combined offline RL using the CQL algorithm with meta-learning using the MAML algorithm to train the UAV using a few shots of experiences stored in an offline static dataset without interacting with the environment in new unseen environments. The proposed algorithm is the only model that converges to the optimum policy through a few SGD steps. The size of the data set and the number of meta-tasks influence the convergence speed, where the convergence is enhanced by increasing the size of the dataset or/and increasing the number of meta-tasks used in training. In addition, the proposed algorithm outperforms traditional schemes, such as DQN, meta-DQN, and CQL, regarding the joint achieved AoI and transmission power. Adapting meta-learning with multi-agent RL (MARL) and assessing its robustness, resiliency, and scalability are open research directions for future works.

\bibliographystyle{IEEEtran}
\bibliography{IEEEabrv,references}
\end{document}